\documentclass[conference]{IEEEtran}
\IEEEoverridecommandlockouts
\usepackage{amsmath,amssymb,amsfonts}
\usepackage{algorithmic}
\usepackage{graphicx}
\usepackage{textcomp}
\usepackage{xcolor}
\def\BibTeX{{\rm B\kern-.05em{\sc i\kern-.025em b}\kern-.08em
    T\kern-.1667em\lower.7ex\hbox{E}\kern-.125emX}}

\usepackage{multirow}
\usepackage{amsmath}
\usepackage{amsfonts}

\usepackage{natbib}

\usepackage{graphicx}
\usepackage{amsmath}
\usepackage{booktabs}
\usepackage{algorithm}
\usepackage{algorithmic}
\newcommand{\agg}{\textrm{agg}}

\begin{document}

\title{Labeled-Graph Generative Adversarial Networks\\
}

\author{\IEEEauthorblockN{Shuangfei Fan}
\IEEEauthorblockA{\textit{Dept.~of Computer Science} \\
\textit{Virginia Tech}\\
sophia23@vt.edu}
\and
\IEEEauthorblockN{Bert Huang}
\IEEEauthorblockA{\textit{Dept.~of Computer Science} \\
\textit{Tufts University}\\
bert@cs.tufts.edu}
}

\maketitle

\begin{abstract}
As a new approach to train generative models, \emph{generative adversarial networks} (GANs) have achieved considerable success in image generation. This framework has also recently been applied to data with graph structures. We propose labeled-graph generative adversarial networks (LGGAN) to train deep generative models for graph-structured data with node labels. We test the approach on various types of graph datasets, such as collections of citation networks and protein graphs. Experiment results show that our model can generate diverse labeled graphs that match the structural characteristics of the training data and outperforms all alternative approaches in quality and generality. To further evaluate the quality of the generated graphs, we use them on a downstream task of graph classification, and the results show that LGGAN can faithfully capture the important aspects of the graph structure.
\end{abstract}


\section{Introduction}
\label{introduction}

Labeled graphs are powerful complex data structures that can describe collections of related objects. Such collections could be atoms forming molecular graphs, users connecting on online social networks, or documents connected by citations. The connected objects, or nodes, may be of different types or classes, and the graphs themselves may belong to particular categories. Methods that reason about this flexible and rich representation can empower analyses of important, complex real-world phenomena. One key approach for reasoning about such graphs is to learn the probability distributions over graph structures. In this paper, we introduce a method that learns generative models for labeled graphs in which the nodes and the graphs may have categorical labels.

A high-quality generative model should be able to synthesize labeled graphs that preserve global structural properties of realistic graphs. Such a tool could be valuable in various settings. One motivating example application is in situations where data owners wish to share graph data but must protect sensitive information. For example, online social network providers may want to enable the scientific community to study the structural aspects of their user networks, but revealing structure could allow reidentification or other privacy-invading inferences \citep{zheleva2012privacy}. A generative model that can create realistic graphs that do not represent real-world users could allow for this kind of study.

Recently \citet{goodfellow2014generative} described generative adversarial networks (GANs), which have been widely explored in computer vision and natural language processing \citep{zhang2017stackgan, yu2017seqgan} for generating realistic images and text, as well as performing tasks such as style transfer. GANs are composed of two neural networks. The first is a generator network that learns to map from a latent space to the distribution of the target data, and the second is a discriminator network that tries to distinguish real data from candidates synthesized by the generator. Those two networks compete with each other during training, and each improves based on feedback from the other. The success of this general GAN framework has proven it to be a powerful tool for learning the distributions of complex data. 

Motivated by the power of GANs, researchers have also used them for generating graphs. \citet{bojchevski2018netgan} proposed NetGAN, which uses the GAN framework to generate random walks on graphs. However, it is unable to generate graphs with node labels, a critical feature of some graph-structured data. \citet{de2018molgan} proposed MolGAN, which generates molecular graphs using the combination of a GAN framework and a reinforcement learning objective. However, our experiments will demonstrate that it lacks the generality to accommodate graphs with different structures and different sizes. 

The rapid development of deep learning techniques has also led to advances in representation learning in graphs. Many works have used deep learning structures to extract high-level features from nodes and their neighborhoods to include both node and structure information \citep{kipf2017semi,fan2020attention,hamilton2017inductive,fan2017recurrent}. These methods have been shown to be useful for many applications, such as link prediction and collective classification. 

Building on these advances, we propose a \emph{labeled-graph generative adversarial network} (LGGAN), a deep generative model trained using a GAN framework to generate graph-structured data with node labels. LGGAN can be used to generate various kinds of graph-structured data, such as citation graphs, knowledge graphs, and protein graphs. Specifically, the generator in an LGGAN generates an adjacency matrix as well as labels for the nodes, and its discriminator uses a graph convolution network \citep{kipf2017semi} with residual connections to identify real graphs using adaptive, structure-aware higher-level graph features. Our approach is the most powerful deep generative method that addresses the generation of labeled graph-structured data. In experiments, we demonstrate that our model can generate realistic graphs that preserve important properties from the training graphs.
We evaluate our model on various datasets with different graph types---such as ego networks and proteins---and with different sizes. Our experiments demonstrate that LGGAN effectively learns distributions of different graph structures and that it can generate realistic large graphs while preserving structural and distributional properties.

\begin{figure}[tb]
\centerline{\includegraphics[width=0.5\linewidth]{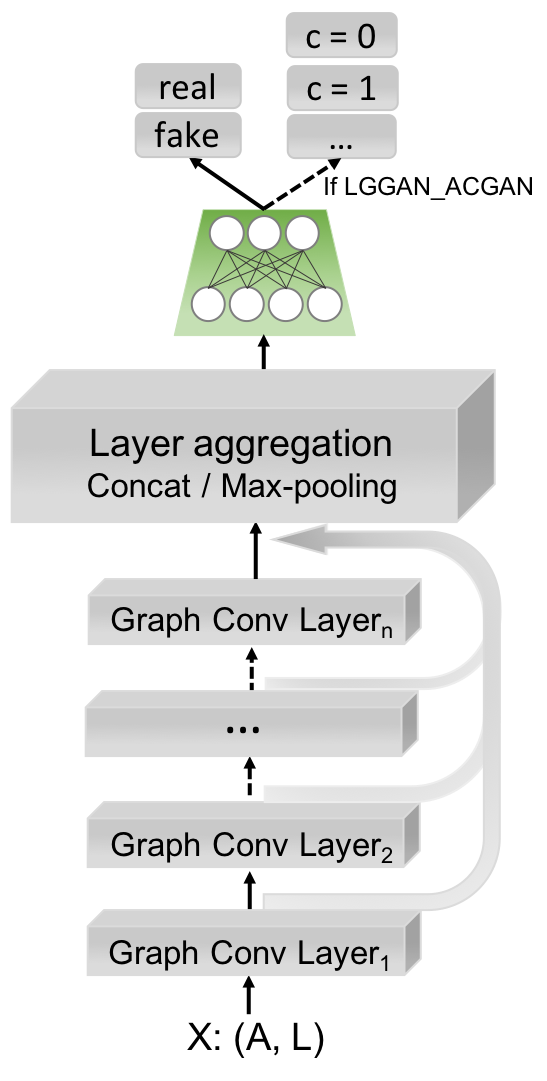}}
\caption{The structure of the LGGAN discriminator with residual connections.}
\label{fig:discriminator}
\end{figure}

\section{Related Work}
\label{related work}

Generative graph models were pioneered by \citet{erdos59a}, who introduced random graphs where each possible edge appears with a fixed independent probability. More realistic models followed, such as the preferential attachment model of \citet{albert2002statistical}, which grows graphs by adding nodes and connecting them to existing nodes with probability proportional to their current degrees. \citet{goldenberg2010survey} proposed the stochastic block model (SBM), and \citet{airoldi2008mixed} proposed the mixed-membership stochastic block model (MMSB). The SBM is a more complex version of the Erd\" os-R\'enyi (E-R) model that can generate graphs with multiple communities. In SBMs, instead of assuming that each pair of nodes has identical probability to connect, they predefine the number of communities in the generated graph and have a probability matrix of connections among different types of nodes. Compared to the E-R model, SBMs are more useful since they can learn more nuanced distributions of graphs from data. However, SBMs are still limited in that they can only generate graphs with this kind of partitioned community structure. 

With the recent development of deep learning, some works have proposed deep models to represent the distribution of graphs. \citet{li2018learning} proposed DeepGMG, which introduced a framework based on graph neural networks. They generate graphs by expansion, adding new structures at each step. \citet{you2018graphrnn} proposed GraphRNN, which decomposes the graph generation into generating node and edge sequences from a hierarchical recurrent neural network. Some other works have explored adopting variational autoencoders (VAEs) to generate graphs, such as the GraphVAE \citep{simonovsky2018graphvae} and Junction Tree VAE \citep{jin2018junction} models. Simultaneously, researchers have also been developing other implicit yet powerful methods for generating graphs, especially based on the success of generative adversarial networks \citep{goodfellow2014generative}. For example, \citet{bojchevski2018netgan} proposed NetGAN, which uses the GAN framework to generate random walks on graphs from which structure can be inferred, and \citet{de2018molgan} proposed MolGAN to generate molecular graphs using the combination of the GAN framework and reinforcement learning.

 However, these recently proposed deep models are either limited to generating small graphs with 40 or fewer nodes \citep{li2018learning} or to generating specific types of graphs such as molecular graphs \citep{you2018graph, de2018molgan} (needing specialized tools to calculate molecule-specific loss). Importantly, most of these recently proposed methods cannot generate labeled graphs.
 
 \section{Model}


In this section, we introduce LGGAN and how it trains deep generative models for graph-structured data with node labels. 

\subsection{Alternative GAN Frameworks}

\begin{figure*}[tb]
\centerline{\includegraphics[width=0.8\linewidth]{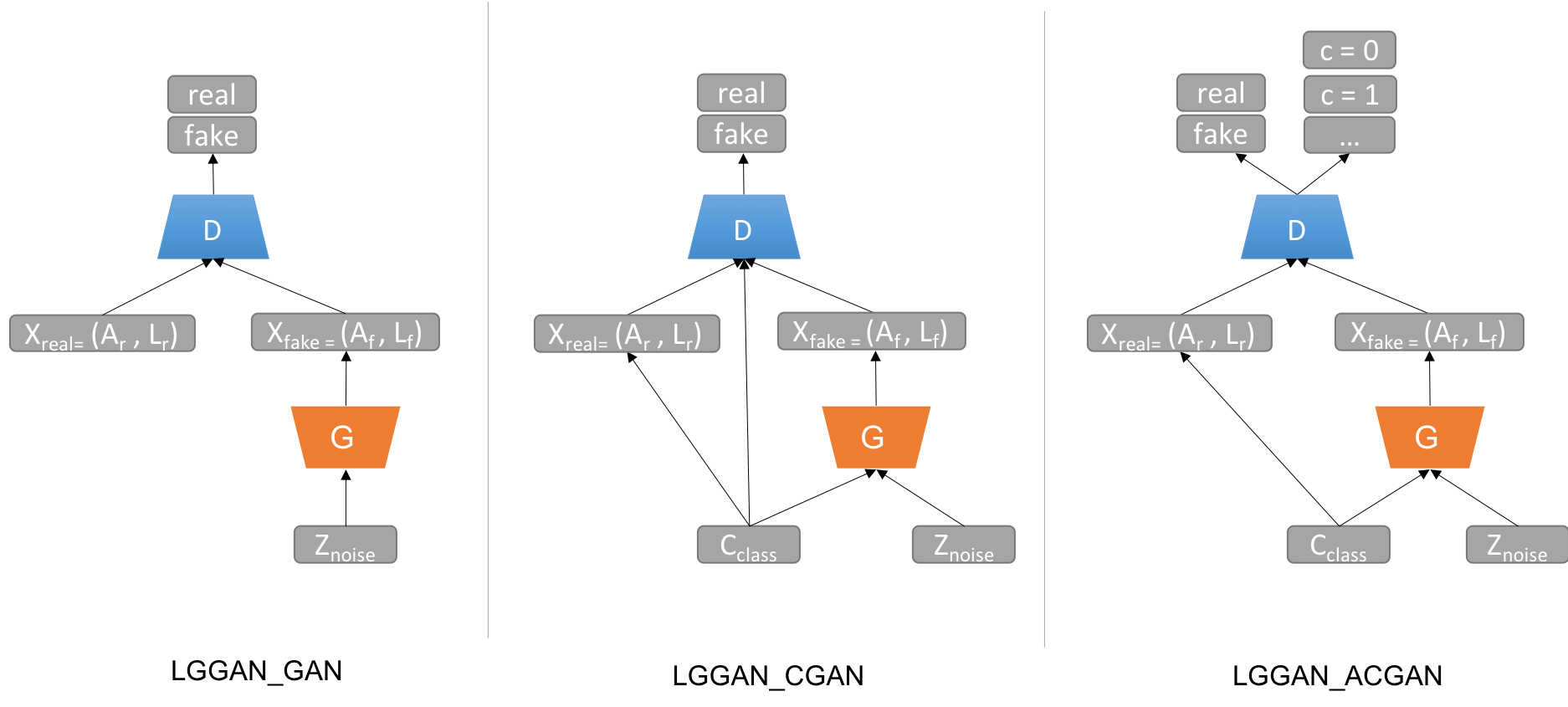}}
\caption{The adversarial training framework of LGGAN using different GAN structures. Conditional GANs and Auxiliary Conditional GANs incorporate increasing amounts of secondary information to aid the training process.}
\label{fig:GAN_framework}
\end{figure*}

Since the GAN framework was introduced by \citet{goodfellow2014generative}, many variations have been proposed that proved to be powerful for generation tasks. Therefore, we adopt three popular variations and compare how well they perform for our task of labeled graph generation. We use the traditional, original GAN approach as the first approach. Beyond the traditional GAN framework, we use two other methods that include extra information: classification labels for the graphs themselves. The first follows the \emph{conditional GAN} framework \citep{mirza2014conditional}, which feeds the graph label as an extra input to the generator in addition to the noise $z$. We can use this label to generate graphs of different types. To improve on this, our last variation uses the \emph{auxiliary conditional GAN} (AC-GAN) framework \citep{odena2017conditional}, in which the discriminator not only distinguishes whether the graph is real or fake, but it also incorporates a classifier of the graph labels. The structure of all those three variations is illustrated in Figure \ref{fig:GAN_framework}.

\textbf{Original GAN~~}
Generative adversarial networks (GANs) \citep{goodfellow2014generative} train implicit generative models by competitively training two neural networks. The first is the generative model $G$, which learns to map from a latent space to the distribution of the target data. The second network is the discriminative model $D$, which tries to separate the real data and the candidates predicted by the generator. These two networks compete with each other during training, via different objective functions. They adapt to improve itself based feedback from each other. The generator $G$ and the discriminator $D$ can be seen as two players in a minimax game where where the generator $G$ tries to produce samples realistic enough to fool the discriminator, and the discriminator $D$ tries to differentiate the samples from the real data and the generator correctly. The objective for GAN training is: 
\begin{equation}
\begin{aligned}
\min_{G}\max_{D} V(D, G) = &\mathbb{E}_{\pmb{x}\sim p_{\textrm{data}}(\pmb{x})}
\left[\log D(\pmb{x})\right]+ \\
&\mathbb{E}_{\pmb{z}\sim p_{\pmb{z}}(\pmb{z})}\left[\log \left(1 - D(G(\pmb{z})\right)\right].
\end{aligned}
\label{equ:GAN objective}
\end{equation}

\textbf{Conditional GAN~~}
After the original GAN was proposed, \citet{mirza2014conditional} introduced a conditional variant, which can be constructed by simply modifying the GAN framework to feed the class label $\pmb{c}$ to both the generator and discriminator. By doing this, the model is then able to generate fake data that are conditioned on class labels. This ability means that this model gives more control over the generated data. The objective function of this conditional GAN is similar to the original version, only adding the condition of $\pmb{c}$ to both generator and discriminator:
\begin{equation}
\begin{aligned}
\min_{G}\max_{D} V(D, G) = &\mathbb{E}_{\pmb{x}\sim p_{\textrm{data}}(\pmb{x})}
\left[\log D(\pmb{x}|\pmb{c})\right]+ \\
&\mathbb{E}_{\pmb{z}\sim p_{\pmb{z}}(\pmb{z})}\left[\log \left(1 - D(G(\pmb{z}|\pmb{c})\right)\right] .
\end{aligned}
\label{equ:C-GAN objective}
\end{equation}


\textbf{AC-GAN~~}
Beyond the conditional GAN,  \citet{odena2017conditional} proposed the AC-GAN variant that extends the idea of conditional GANs. In the AC-GAN framework, each generated sample has a pre-defined (usually randomly chosen) class label, 
$\pmb{c}\sim p_{\pmb{c}}$ in addition to the noise $\pmb{z}$ as the input to the generator $G$. The generator $G$ then uses both to generate samples $X_{\textit{fake}} = G(\pmb{c}, \pmb{z})$. The discriminator will output probability distributions over both data and the class labels, $P(S | X), P(C | X) = D(X)$. The objective function of AC-GAN has two parts: the log-likelihood of the correct data, $L_S$, and the log-likelihood of the correct class, $L_C$.
\begin{equation}
\begin{aligned}
L_S = & \mathbb{E} \left[ \log P(S = \textit{real} | X_{\textit{real}})\right]\\
& +\mathbb{E}[\log P(S = \textit{fake} | X_{\textit{fake}})],
\end{aligned}
\label{eq:ls}
\end{equation}

\begin{equation}
\begin{aligned}
L_C = & \mathbb{E}[ \log P(C = c | X_{\textit{real}})]\\
& +\mathbb{E}[\log P(C = c | X_{\textit{fake}})].
\end{aligned}
\label{eq:AC-GAN objective}
\end{equation}

\subsection{Architecture}
\label{architecture}

In this section, we provide details on the LGGAN architecture. As in the standard GAN framework, LGGAN consists of two main components: a generator $G$ and a discriminator $D$. The generator $G$ takes a sample from a prior distribution and generates a labeled graph $g$ represented by an adjacency matrix $A$ and a node label matrix $L$. The discriminator $D$ then trains to distinguish samples from the dataset and samples from the generator. In LGGAN, both the generator and the discriminator are trained using CT-GAN \citep{WeiGL0W18}, an improved version based on the Wasserstein GAN approach \citep{gulrajani2017improved}.

\textbf{Generator~~} LGGAN's generative model uses a multi-layer perceptron (MLP) to produce the graph. The generator $G$ takes a random vector $z$ sampled from a standard normal distribution and outputs two matrices: (1) $L \in R^{N \times C}$, which is a one-hot vector that defines the node labels, and (2) the adjacency matrix $A \in R^{N \times N}$, which defines the connections among nodes in graphs. The architecture uses a fixed \emph{maximum} number of nodes $N$, but it is capable of generating structures of fewer nodes by dropping the nodes that are not connected to any of the other node in the generated graph. Since both the adjacency matrix $A$ and the label matrix $L$ are discrete and the categorical sampling procedure is non-differentiable, we directly use the continuous objects $A$ and $L$ during the training procedure. The original GAN structure uses the Jensen-Shannon (JS) divergence to measure the distance, which then cannot be used to generate discrete data. Therefore, we use variants of GANs that are based on Wasserstein distance, such as CT-GAN \citep{WeiGL0W18} and WGAN-GP \citep{gulrajani2017improved}, which have been shown to be applicable to discrete data such as text. 

We use an MLP to directly generate the adjacency matrix because it circumvents many challenges with alternate approaches that sequentially generate edge structures, e.g., \citep{you2018graphrnn}. Because there is a direct path from the latent representation to each edge appearance variable, gradients are less prone to vanishing as can happen with sequence models. The computational cost is a concern, but sequential methods also iteratively consider all possible edges, incurring the same quadratic cost as our MLP generator. 

\begin{table*}[tbhp]
  \caption{Details of the graph datasets.}

  \label{tab:dataset}
  \centering
  \scalebox{1.1}{
  \begin{tabular}{ccccccc}
    \toprule
     Graph Types & Datasets     & \# Graphs  &  \# Graph classes & Avg. $\vert V\vert$ & Avg. $\vert E\vert$ & \# Node labels\\
    \hline  
    \multirow{4}{*}{Citation graphs} & Cora-small & 512 & 7  &  38.7 & 61.6 &7 \\
    & CiteSeer-small     & 512 & 6  &  44.2  & 82.7 & 6\\
    & Cora-large & 128  & 7 & 175.3 & 326.3 & 7\\
    & CiteSeer-large     & 128 & 6  &  172.5  &414.7 & 6 \\
    \hline  
    \multirow{2}{*}{Protein graphs}  & PROTEINS & 384 & 2 & 28.1 &53.4 & 3 \\
    & ENZYMES & 256 & 6 & 39.4 &77.7 & 3 \\
    \bottomrule
  \end{tabular}
  }
\end{table*}

\textbf{Discriminator~~} The discriminator $D$ takes a graph sample as input, represented by an adjacency matrix $A$ and a node label matrix $L$, and it outputs a scalar value and a one-hot vector for the class label. For the discriminator, we use a graph convolutional network (GCN) \citep{kipf2017semi}, which is a powerful neural network architecture for representation learning with complex graph structures. GCNs propagate information along graph edges with graph convolution layers. GCNs are also permutation-invariant, which is important when analyzing graphs because they are usually considered unordered objects. 


\textbf{Residual Connections~~}
As reported in by \citet{kipf2017semi} when they introduced graph convolutional networks, one issue that affect the application of GCN on complex graphs is that it lacks the ability to train with deeper models. Our early stage experiment results show that the performance of GCN stops improving with more than 3 GC layers.

To deal with this problem and unblock the applications of GCN on graphs with more complicated structures, we add residual connections \citep{he2016deep} between hidden layers of the GCN to allow the model to fuse information from previous layers. We find that these residual connections circumvent the issues reported by \citet{kipf2017semi} that limit their GCNs to only two or three layers. Allowing more depth is important because some graph types, such as proteins and molecules, have complex structure that requires incorporation of information from nodes further in graph distance than are reachable with only three graph convolutions.

With residual connections, each layer of our GCN discriminator propagates with the following rule:
\begin{equation}
\begin{aligned}
H^{(l+1)} = \sigma \left(\tilde{D}^{-\frac{1}{2}}\tilde{A}\tilde{D}^{-\frac{1}{2}}H^{(l)}W^{(l)}\right),
\end{aligned}
\label{equ:GCN1}
\end{equation}
where $H(l) \in R^{N\times D}$ is the output matrix at the $l-1$th layer, $I_N$ is the identity matrix, $\tilde{A} = A + I_N$ is the adjacency matrix of the graph $g$ with self-connections added, $\tilde{D}_{ii} = \sum_j \tilde{A}_{ij}$ 
is the diagonal degree matrix of graph $g$, $W^{(l)} \in R^{D\times F}$ is the trainable weight matrix at the $l$th layer, and $\sigma(\cdot)$ denotes an activation function (such as the sigmoid or ReLU \citep{nair2010rectified}). Since we do not include node attributes, we set $H(0) = I_N$, where $I_N$ is the identity matrix.

After $n$ layers of propagation via graph convolutions, we aggregate the outputs from each layer with an aggregation function $\agg$, such as concatenation and max-pooling. We then concatenate the aggregated matrix with the node label matrix $L$ and output $Z_g$ as the final representation we learned for graph $g$:
\begin{equation}
\begin{aligned}
Z_g &= f(X, A, L) 
 = \left[\agg\left(H^{(1)},\dots, H^{(n)}\right) ; L\right].
\end{aligned}
\label{equ:GCN2}
\end{equation}
%
The representation $Z_g$ of the graph will be further processed by a linear layer to produce the outputs of the discriminator: the graph-level scalar probability of the input being real data and a classifier to predict the category that the graph belongs to with a one-hot vector $c$. We illustrate the structure of this discriminative model in Figure \ref{fig:discriminator}.

In Section \ref{sec:residual}, we evaluate the influence of the depth of GCN with and without residual connections and also with different aggregate functions to guide how to choose from different settings of LGGAN for the experiments.


\subsection{Training}

GANs \citep{goodfellow2014generative} train via a min-max game with two players competing to improve themselves. In theory, the method converges when it reaches a Nash equilibrium, where the samples produced by the generator match the data distribution. However, this process is highly unstable and often results in problems such as mode collapse \citep{goodfellow2016nips}. To deal with the most common problems in training GANs, such as mode collapse and unstable training, we use the CT-GAN \citep{WeiGL0W18} framework, which is a state-of-the-art approach. CT-GAN adds a consistency term to the Wasserstein GANs (WGAN-GP) \citep{salimans2016improved} that preserves Lipschitz continuity in the training procedure of WGAN-GPs. We also adopt several techniques such as feature matching and minibatch discrimination that were shown to encourage convergence and help avoid mode collapse \citep{salimans2016improved}.  



\subsection{Node Ordering}

A common representation for graph structure uses adjacency matrices. However, using matrices to train a generative model introduces the issue of how to define the node ordering in the adjacency matrix. There are $n!$ permutations of $n$ nodes, and it is time consuming to train over all of them.

For LGGAN, we use the GCN framework with residual connections and a node aggregation operator as the discriminator. This discriminator is therefore invariant to node ordering, avoiding the issue. However, for the generator, we use an MLP, which does depend on node ordering. Therefore, we adapt the approach by \citet{you2018graphrnn} and arrange the nodes in a breadth-first-search (BFS) ordering for each training graph.

In particular, we preprocess the adjacency matrix $A$ and node label matrix $L$ by feeding them into a BFS function. This function takes a random permutation $\pi_g$ of the nodes in graph $g$ as input, picks a node $v_i$ as the starting node, and then outputs another permutation $\pi_g'$ that is a BFS ordering of the nodes in graph $g$ starting from node $v_i$. By specifying a structure-determined node ordering for the graph, we only need to consider all possible BFS orderings, rather than all possible node permutations. This reduction makes a significant difference for computational complexity when graphs are large.

\begin{figure*}[tb]
\centerline{\includegraphics[width=1.0\linewidth]{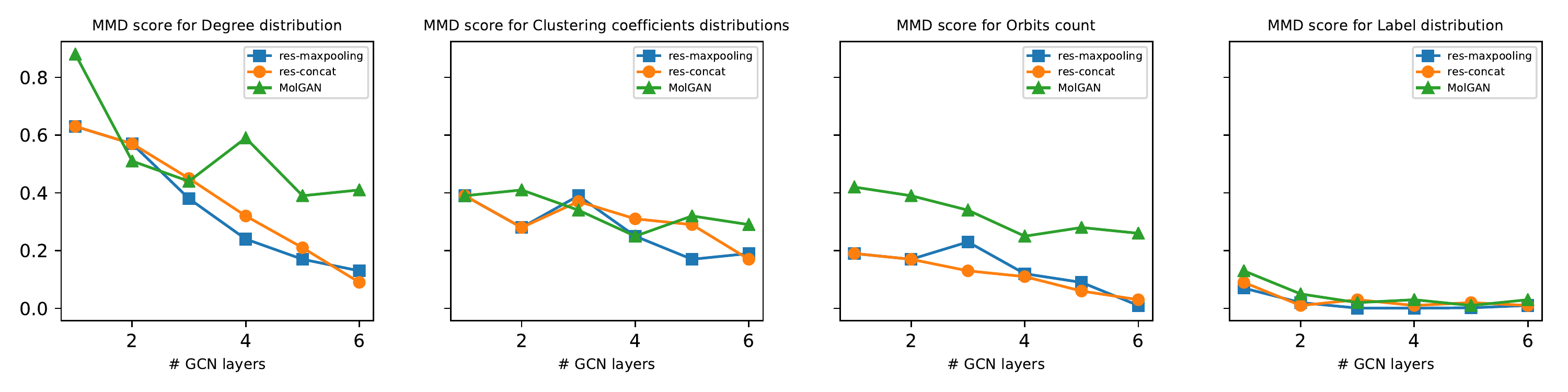}}
\caption{Comparison of the results with different GCN layers between MolGAN and LGGAN with different aggregate functions.}
\label{fig:res_net}
\end{figure*}

\section{Experiments}


In this section, we compare LGGAN with other graph generation methods to demonstrate its ability to generate high-quality labeled graphs in diverse settings. We further evaluate the quality of the generated graphs by applying LGGAN to a downstream task of graph classification.

\subsection{Compared Methods}


 We compare our model against various traditional generative models for graphs, as well as some recently proposed deep graph generative models. Regarding the baselines with traditional models, we compare against the Erd\" os-R\'enyi (E-R) model \citep{erdos59a}, Barab\'asi-Albert (B-A) model \citep{albert2002statistical}, and mixed-membership stochastic block (MMSB) model \citep{airoldi2008mixed}. Moreover, we also compare with some recently proposed deep graph generative models such as DeepGMG \citep{li2018learning}, GraphRNN \citep{you2018graphrnn} and MolGAN  \citep{de2018molgan}. Notice that among them, few current approaches are designed to generate labeled graphs. One exception is MolGAN, which is designed to generate molecular graphs and needs specialized evaluation methods specific to that task. Therefore, in order to compare with it, we drop the molecule-specific reward network and the corresponding RL loss to apply it to other types of graphs, please refer to Section \ref{sec:residual} for more details about the corresponding experiment results and discussions regarding the comparison between MolGAN and LGGAN.

\subsection{Datasets}
\label{dataset}

We perform experiments on different kinds of datasets with varying sizes and characteristics. Details about the statistics for these datasets are listed in Table \ref{tab:dataset}.



 

\textbf{Citation graphs~~} We test on scientific citation networks, using the Cora and CiteSeer datasets \citep{sen:aimag08}. The Cora dataset is a collection of 2,708 machine learning publications categorized into seven classes, and the CiteSeer dataset is a collection of 3,312 research publications crawled from the CiteSeer repository. To test the ability of LGGAN to generate larger graphs, we extract different subsets with different graph sizes by constraining the number of nodes in graph $|V|$. For small datasets (denoted Cora-small and CiteSeer-small), we extract two-hop and three-hop ego networks with $30 \leq |V| \leq 50$. For the large datasets (denoted Cora and CiteSeer), we extract three-hop ego networks with $150 \leq |V| \leq 200$. We set the graph label to be the node label of the center node in the ego network.

\textbf{Protein graphs~~} We also test on multiple collections of protein graphs. The ENZYMES dataset consists of 600 enzymes \citep{schomburg2004brenda}. Each enzyme is labeled with one of the six enzyme commission (EC) code top-level classes. The PROTEINS dataset includes proteins from the dataset of enzymes and non-enzymes created by \citet{dobson2003distinguishing}. The graphs are categorized as enzymes or non-enzymes. 

\begin{table*}[tb]
  \caption{Comparison of LGGAN and other generative models on different graph-structured data using MMD evaluation metrics.}
  \label{tab:evaluation}
  \centering
  \scalebox{1.0}{
  \begin{tabular}{ccccccccccccc}
    \toprule
    & \multicolumn{4}{c}{Cora-small} & \multicolumn{4}{c}{CiteSeer-small} & \multicolumn{4}{c}{Cora} \\
    \cmidrule(lr){2-5}
    \cmidrule(l){6-9}
    \cmidrule(l){10-13}

      &  Degree  &   Clustering &  Orbit &  Label  & Degree  &   Clustering  &  Orbit &  Label &  Degree  &   Clustering &  Orbit &  Label    \\

    \cmidrule(lr){1-13}
    E-R & 0.68  & 0.94 & 0.48  & N/A  & 0.63  & 0.86 & 0.12 & N/A & 0.88  & 1.45  & 0.27& N/A  \\
    B-A     & 0.31 & 0.53 & 0.11 & N/A & 0.37  & 0.18 & 0.11 & N/A  & 0.54 & 1.06 &0.16 & N/A     \\
    MMSB     & 0.21 & 0.68 & 0.07 & 0.48 & {\bf 0.17}  & 0.50 & 0.11 &  0.32   & {\bf 0.12}     & 0.68 &0.09&  0.49  \\
    DeepGMG &  0.34 & 0.44 & 0.27 & N/A  &  0.27 & 0.36 &0.20 & N/A  & -& -& -&-    \\
    GraphRNN & 0.26 & 0.38  &0.39 & N/A & 0.19 & 0.20 & 0.39& N/A  &  0.20&  0.46& 0.11&N/A  \\
    LGGAN &  {\bf 0.13}& {\bf 0.08} &{\bf 0.03} &{\bf 0.11} &{\bf 0.17}& {\bf 0.13} & {\bf 0.04}&{\bf 0.09}  &  0.15& {\bf 0.21}&{\bf 0.06}&{\bf 0.01}  \\
    \bottomrule
  \end{tabular}
  }
  \scalebox{1.0}{
  \begin{tabular}{ccccccccccccc}
     & \multicolumn{4}{c}{PROTEINS} & \multicolumn{4}{c}{ENZYMES}& \multicolumn{4}{c}{CiteSeer} \\
    \cmidrule(lr){2-5}
    \cmidrule(l){6-9}
    \cmidrule(l){10-13}


      & Degree  &  Clustering & Orbit & Label  & Degree  &  Clustering  & Orbit & Label & Degree  &  Clustering & Orbit & Label  \\
    \cmidrule(lr){1-13}
    E-R  & 0.31  & 1.06  & 0.28& N/A & 0.38  & 1.26   &0.08 & N/A & 0.82  & 1.57   &{\bf 0.06} & N/A \\
    B-A          & 0.93 & 0.88 &0.05 & N/A  & 1.17 & 1.08 & 0.51& N/A & 0.32 & 1.04 & 0.08& N/A \\
    MMSB     & 0.46     & 1.05 &0.21&  {\bf 0.01} & 0.55  & 1.08 &0.05&  0.92 & {\bf 0.08}  & 0.50 &0.11&  0.32\\
    DeepGMG   & 0.96& 0.63& 0.16&N/A & 0.43&  0.38& 0.08&N/A& -& -& -&-\\
    GraphRNN   &  {\bf 0.04}&  0.18& 0.06&N/A & {\bf 0.06}&  0.20& 0.07&N/A & 0.20&  1.15& 0.14&N/A\\
    LGGAN  &  0.18& {\bf 0.15}&{\bf 0.02}&{\bf 0.01} &  0.09&  {\bf0.17}&{\bf0.03}&{\bf 0.01}&  0.25 &  {\bf0.12}&{\bf0.06}&{\bf 0.15}\\
    \bottomrule
  \end{tabular}
  }
\end{table*}

\subsection{Metrics for the Quality of Generated Labeled Graphs}
\label{sec:evaluation}

To evaluate the quality of the generated graphs, we follow the approach used by \citet{you2018graphrnn}: We compare a distribution of generated graphs with that of real ones by measuring the \emph{maximum mean discrepancy} (MMD) \citep{gretton2012kernel} of graph statistics, capturing how close their distributions are.
We use four graph statistics to evaluate the generated graphs: degree distribution, clustering coefficient distribution, node-label distribution, and average orbit count statistics.


\subsection{Evaluation}
To explore the power of LGGAN, we test it with many variants and compare their advantages based on (1) three different GAN frameworks (vanilla GAN \citep{goodfellow2014generative}, conditional GAN \citep{mirza2014conditional} and AC-GAN \citep{odena2017conditional}) and (2) two different discriminative models (a simple multi-layered perceptron (MLP) and a GCN with residual connections). 

\subsection{Comparing Different Variations of LGGAN}
\label{comparegans}

LGGAN is a flexible framework that can adopt different settings for each part of the training setup, such as the GAN framework and the model for generator and discriminator. The GAN variations we consider are (1) LGGAN\_GAN, which uses the original GAN framework \citep{goodfellow2014generative}, (2) LGGAN\_CGAN, which uses the conditional GAN framework \citep{mirza2014conditional}, and (3) LGGAN\_ACGAN, which uses the AC-GAN \citep{odena2017conditional} framework.

\begin{table*}[tb]
\caption{Comparison of LGGAN with different GAN frameworks and discriminative models on Cora-small and ENZYMES datasets.  }
\label{tab:compare}
  \centering
  \scalebox{1.1}{
  \begin{tabular}{lcccccccc}
  \toprule
    \multirow{2}{*}{GAN frameworks}   &\multicolumn{4}{c}{Cora-small} &  \multicolumn{4}{c}{ENZYMES} \\ 

    \cmidrule(lr){2-5}
    \cmidrule(l){6-9}
     & Degree  &  Clustering & Orbit  & Label & Degree  &  Clustering & Orbit  & Label\\
      \cmidrule(lr){1-9}

    LGGAN\_GAN\_s & 0.27  &  0.18 &0.03 & 0.37 & 0.67  &  0.88 & 0.004& 0.01\\
    LGGAN\_GAN & 0.21  &  0.14 &{\bf 0.01} & 0.15 &0.31  &  0.20 & {\bf 0.01}& {\bf 0.01}  \\
    LGGAN\_CGAN\_s & 0.18  &  0.18 &0.006 & 0.35 & 0.53  &  0.69 &0.04 & 0.004\\
    LGGAN\_CGAN &{\bf 0.10}  &  0.24 & {\bf 0.01}& 0.19& 0.23  &  {\bf 0.13} &0.02 & {\bf 0.01} \\
    LGGAN\_ACGAN\_s & 0.14  &  {\bf 0.009} & 0.06 & 0.13 & 0.51  &  0.29 & 0.03 & 0.01\\
    LGGAN\_ACGAN & 0.13  &  {\bf 0.08} &  0.03 & {\bf 0.11}&{\bf 0.09}  &  0.17 & {\bf 0.01} & {\bf 0.01} \\
    \bottomrule
  \end{tabular}
  }
\end{table*}

We evaluate these three possible setups for LGGAN on different graph-structured data to measure the quality of the learned generative models. We run experiments on two datasets: Cora-small and ENZYMES. The results are listed in Table \ref{tab:compare}. Among all the three GAN frameworks, LGGAN\_ACGAN achieves the best results on both datasets regardless of which discriminative model is used. This result matches with our expectations, since the AC-GAN framework incorporates the class information allowing it to learn a better embedding and to propagate that information to the generator. Therefore, in the following experiments, we choose to use the AC-GAN framework as the primary setting for LGGAN.

\subsection{Evaluating the Design of the Residual GCN Discriminator}
\label{sec:residual}


We evaluate the influence of residual connections and the depth of the GCN discriminator. We report the results for graph generation on the ENZYMES dataset based on the four evaluation metrics mentioned in Section \ref{sec:evaluation}. Through these tests, we aim to investigate the following design aspects: (1) the GCN depth, (2) different aggregate functions (i.e., max-pooling and concatenation), and (3) residual connections. To evaluate the effectiveness of residual connections, we compare LGGAN to another model, MolGAN (without the reward network due to specialization). It used a similar framework for generating molecules but suffered from the limitations of GCN with shallow architectures. The results are plotted in Figure \ref{fig:res_net}.

In those plots, the green line represents the results using the MolGAN model without the reward network (due to specialization). Compared to MolGAN, one of the main differences is that we introduced the residual connections into the discriminator of LGGAN to help mitigate the influence of vanishing gradient problem in deep graph convolutional networks. According to the plots, the performance of MolGAN does not have noticeable improvement with more than two or three graph convolutional layers. However, with LGGAN, when adding the residual connections, using either aggregate function can train deeper GCNs with more than five or six layers, achieving high quality results that outperform other models we compare to in Section \ref{sec:compare_baseline}.  There is no notable difference between the aggregate functions. Since max-pooling does not introduce any additional parameters to learn, we use max-pooling as the aggregation function for residual connections in the remaining experiments.  

To further evaluate the performance of the GCN-based discriminative model, we also compare LGGAN with a simple version where we use MLP as the discriminator.

\section{Comparing Different Models for the Discriminator}

For the discriminator, We also compare two different discriminative models, a simple multi-layered perceptron (MLP) and GCN with residual connections and the advanced model is what we proposed in our paper which is a GCN with residual connections. The advanced model is comprised by a series of graph convolution layers and a layer aggregation operator to integrate useful information from each layer for learning more powerful graph representations. We refer to the whole framework using the simple model as LGGAN\_s and using what we proposed as simply LGGAN. 

We evaluate these six different variants for LGGAN (three different GAN frameworks for either LGGAN or LGGAN\_s) on different graph-structured data to measure the quality of the learned generative models. We run experiments on two datasets: cora\_small and ENZYMES. The results are listed in Table \ref{tab:compare}.

Among all the three GAN frameworks, LGGAN\_ACGAN achieves the best results on both datasets regardless of which discriminative model is used. This result matches with our expectations, since the AC-GAN framework incorporates the class information allowing it to learn a better embedding and to propagate that information to the generator.

Also, we noticed that using GCN with residual connections added can improve the quality of generated graphs regardless of which GAN framework is used. An interesting point is that there is a gap between LGGAN\_s and LGGAN on the ENZYMES dataset that is much larger than on the citation networks. This difference may be because the Cora\_small dataset is composed of many small two-hop and three-hop ego networks where the structure is quite simple and uniform---so it could be easier to learn. However, with the ENZYMES dataset, the structure is more complicated and diverse. Therefore, this result reveals that the LGGAN\_s is unable to generalize to complex data. In contrast, the quality of generated graphs with LGGAN using GCN with residual connection is more consistent among different datasets, which suggests that LGGAN can adaptively adjust to different graph-structured data. Based on these results, we use LGGAN\_ACGAN in the remaining experiments and refer to it as LGGAN when comparing with other baselines.

\begin{table*}[tb]
\caption{Comparison of graph classification accuracy with different kernels: graphlet kernel (GK), shortest-path kernel (SP), and Weisfeiler-Lehman subtree kernel (WL) on citation and protein datasets with the other labeled graph generation model MMSB. }
  \centering
  \scalebox{1.2}{
  \begin{tabular}{cccccccccc}
  \toprule
     &\multicolumn{3}{c}{Cora-small} &  \multicolumn{3}{c}{ENZYMES} &  \multicolumn{3}{c}{PROTEINS}\\ 
       \cmidrule(lr){2-4}
       \cmidrule(l){5-7}
       \cmidrule(l){8-10}

    & GK & SP & WL &  GK & SP &  WL &  GK & SP  & WL \\ 

    \cmidrule(l){1-10}
 

    MMSB   & 22.62  &  21.34  & 23.15 & 15.38  &  14.29 &17.86 & 64.32 & 65.61 & 64.29 \\
    LGGAN  &23.44  &26.56 &  26.92  &  23.08 & 26.92 & 30.19&65.61 &66.54 & 69.23  \\
    Real graphs  & 26.56  &  29.69 &34.32  & 23.08   & 29.68 & 35.71 & 69.05 & 73.81 & 76.19 \\
    \bottomrule
  \end{tabular}
  }
 \label{tab:graph_classification}

\end{table*}

\subsection{Evaluating Graph Structure Distributions}
\label{sec:compare_baseline}

We compare LGGAN to other methods for generating graphs---traditional generative models such as E-R, B-A, and MMSB, as well as deep generative models that were proposed recently, such as GraphRNN and DeepGMG. DeepGMG cannot be used to generate large graphs due to its high computational complexity, so the results of DeepGMG on large graph datasets are not available. For each method, we measure two aspects. The first is the quality of the generated graphs, which should be able to mimic typical topology of the training graphs. The second is the generality, where a good generative model should be able to apply to different and complex graph-structured data. Moreover, regarding generality, we want the model to be able to scale up to generate larger networks instead of being restricted to relatively small graphs.

Table \ref{tab:evaluation} lists results from our comparison. LGGAN achieves the best performance on all datasets, with a $90\%$ decrease of MMD on average compared with traditional baselines, and a $30\%$ decrease of MMD compared with the state-of-the-art deep learning method GraphRNN. Although GraphRNN performs well on the two smaller protein-related datasets, ENZYMES and PROTEINS, it does not maintain the same performance on large datasets, such as Cora and CiteSeer. The results of GraphRNN for some datasets are not exactly the same as those reported by \citet{you2018graphrnn} since we are training on graphs extracted from the dataset with different methods and settings.




\subsection{Evaluating Labeled Graphs}

Since we are generating labeled graphs, we also want to evaluate the graph distribution in each class. To do this, we extract subgraphs for each class from both the training graphs and generated graphs and evaluate based on these three metrics (excluding the label distribution). These per-label tests help measure whether the model simply assigns the class based on the label distribution without considering the underlying graph structure. Since among existing methods, only MMSB can be used to directly generate labels, we compare LGGAN to it using the ENZYMES dataset. The results are listed in Table \ref{tab:perclass_eval}. LGGAN can learn a good distribution of the labels, and it is also able to learn the structure within each class much more reliably than the MMSB model.


\begin{table*}[tb]
\caption{Comparison of LGGAN with MMSB on both the graph statistics and average sub-graph statistics of different classes using MMD evaluation metrics on the ENZYMES dataset. }
\label{tab:perclass_eval}
  \centering
  \scalebox{1.2}{
  \begin{tabular}{cccccccc}
  \toprule
  &\multicolumn{4}{c}{Graph statistics} &  \multicolumn{3}{c}{Sub-graph statistics} \\ 

    \cmidrule(lr){2-5}
    \cmidrule(l){6-8}
     & \textbf{D}egree  &   \textbf{C}lustering  &   \textbf{O}rbit  & \textbf{L}abel & Avg. \textbf{D}  &   Avg. \textbf{C} &   Avg. \textbf{O}   \\
    \cmidrule(lr){1-8}
 
    MMSB & 0.55  &  1.08 &0.05 & 0.92& 0.14 & 0.20  &  0.03 \\
    LGGAN & {\bf 0.09}  &  {\bf 0.17} &{\bf 0.03} & {\bf 0.01} &{\bf 0.13}&{\bf 0.15}  &  {\bf 0.01} \\
    \bottomrule
  \end{tabular}
  }
\end{table*}

\subsection{Downstream Task: Graph Classification}
\label{sec:graph_classification}

To further evaluate the quality of LGGAN's generated graphs, we extract the generated examples and use them for a downstream task. We use the synthetic graphs to train a model for graph classification. We first compute a kernel matrix $K \in \mathbb{R}^{n*n}$ for each of a set of graph kernels, where $K_{ij}$ represents an inner product between representations of $G_i$ and $G_j$. Then we can train kernel support vector machines (SVM) to classify the graphs. In our experiment, we choose three popular graph kernels: the graphlet kernel (GK) based on subgraph patterns, the shortest-path kernel (SP) based on random walks, and the Weisfeiler-Lehman subtree kernel (WL) based on subtrees. 

We compare performance when training on synthetic graphs from LGGAN, MMSB (the only compared model that can generate labeled graphs), and real graphs. We run this procedure with three datasets: Cora-small, ENZYMES, and PROTEINS. For each dataset, we average accuracy over ten trials.

We list the results in Table \ref{tab:graph_classification}. 
The accuracy of models trained with graphs generated by LGGAN is close to those trained using the real graphs. In contrast, the models trained with graphs generated by MMSB are significantly less accurate. These results suggest that the graphs generated by LGGAN can better capture the important aspects of graph structure information, which is much similar to real graphs compared to other models, and these results in another way strongly validate that our model is successful.

\begin{figure*}[tbp]
\centering
\begin{minipage}[]{0.495\textwidth}
\begin{center}
\includegraphics[width=1\textwidth]{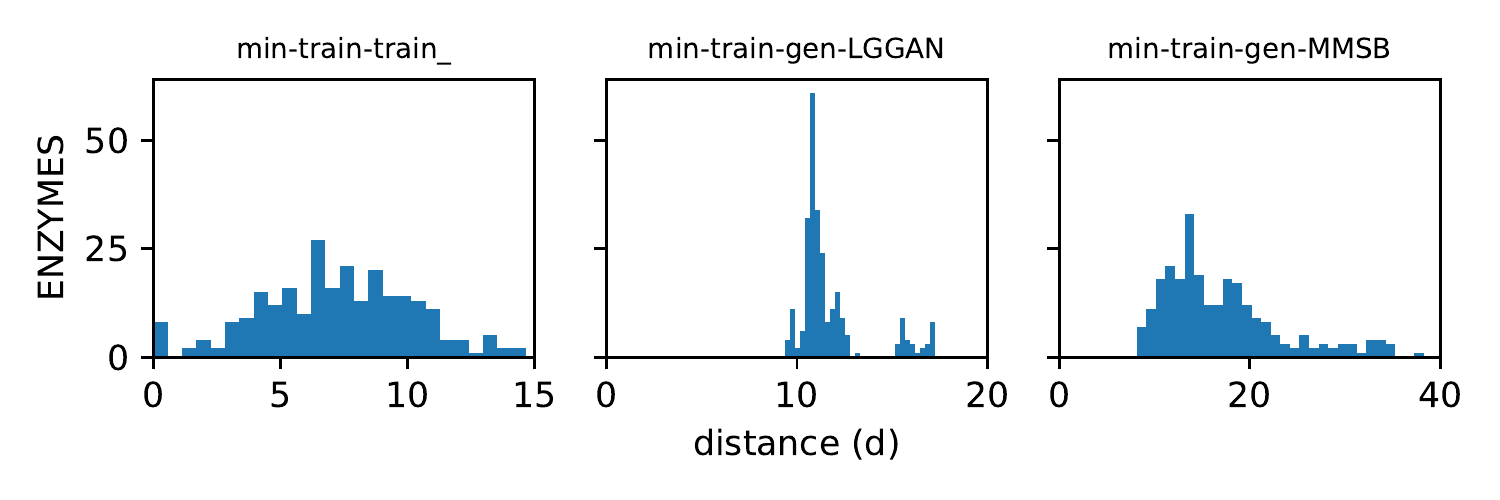}\\
\includegraphics[width=1\textwidth]{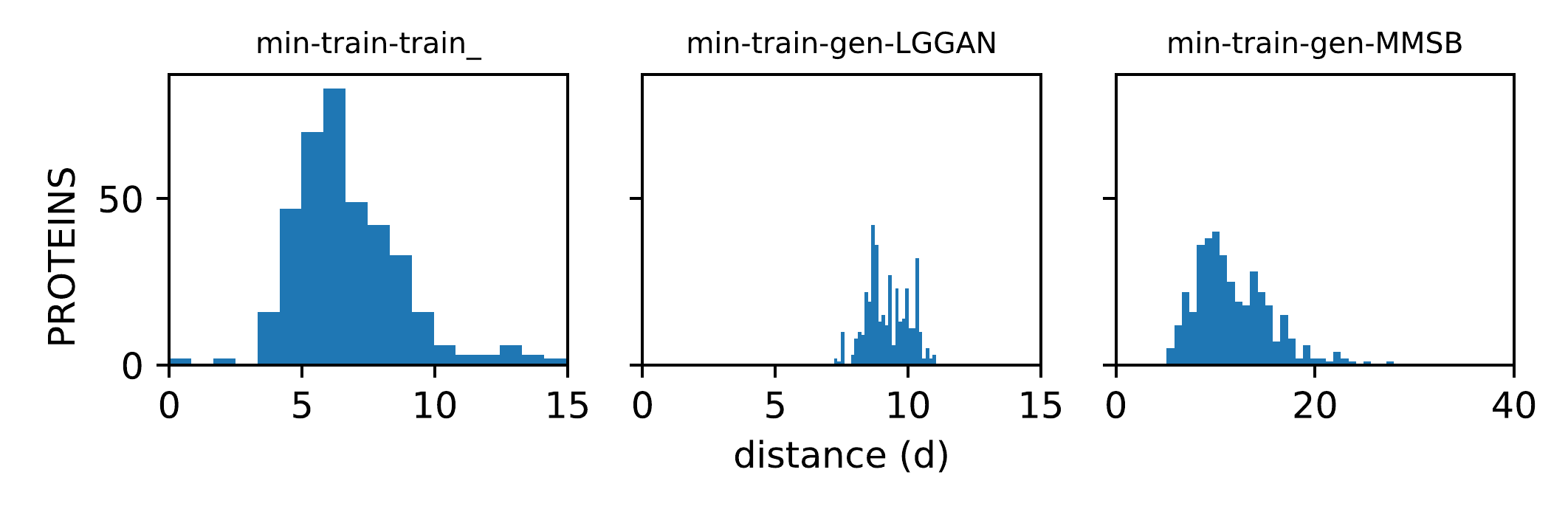} 
\end{center}
\end{minipage}
\begin{minipage}[]{0.495\textwidth}
\begin{center}
\includegraphics[width=1\textwidth]{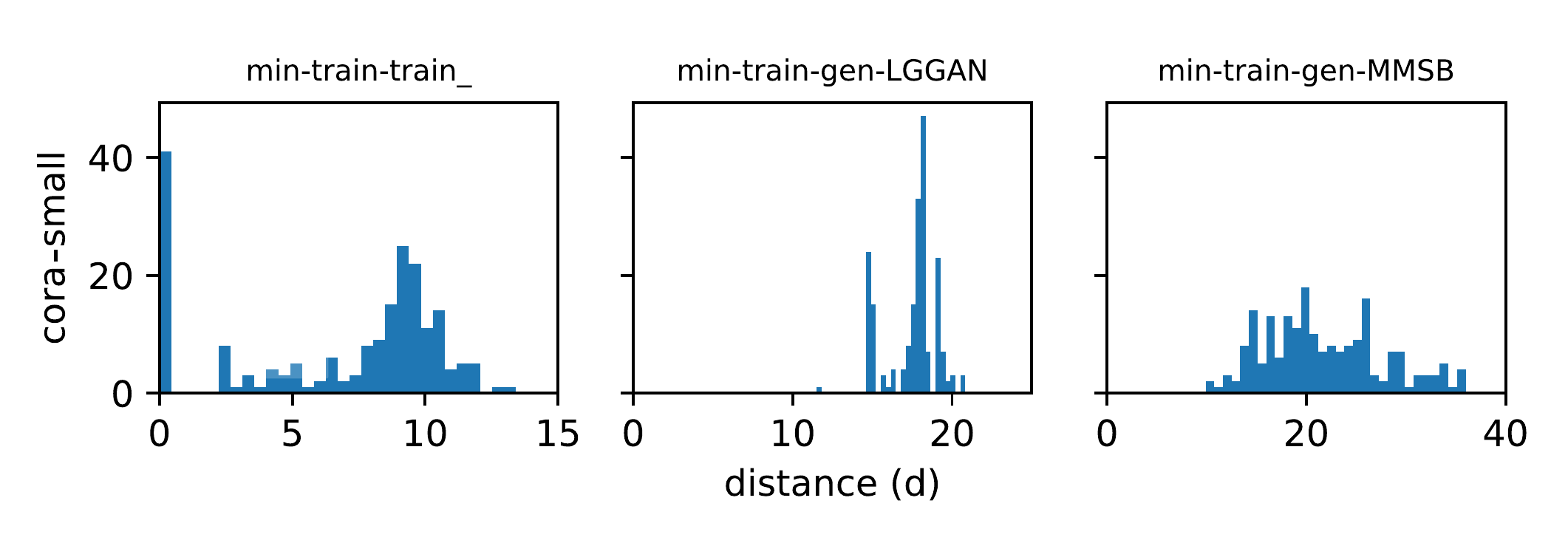} \\
\includegraphics[width=1\textwidth]{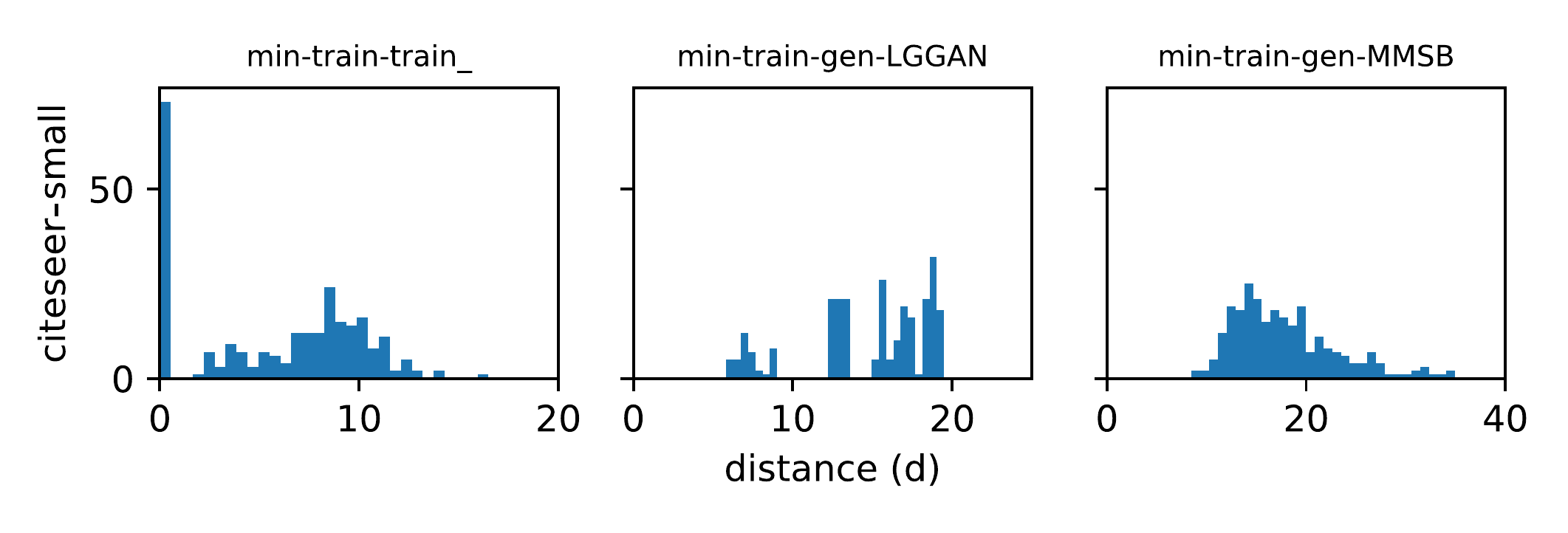} 
\end{center}
\end{minipage}

\caption{Histogram of the distances between training graphs and graphs generated by LGGAN and MMSB.}
\label{fig:hist}
\end{figure*}

\begin{figure*}[tb]
\centerline{\includegraphics[width=0.95\linewidth]{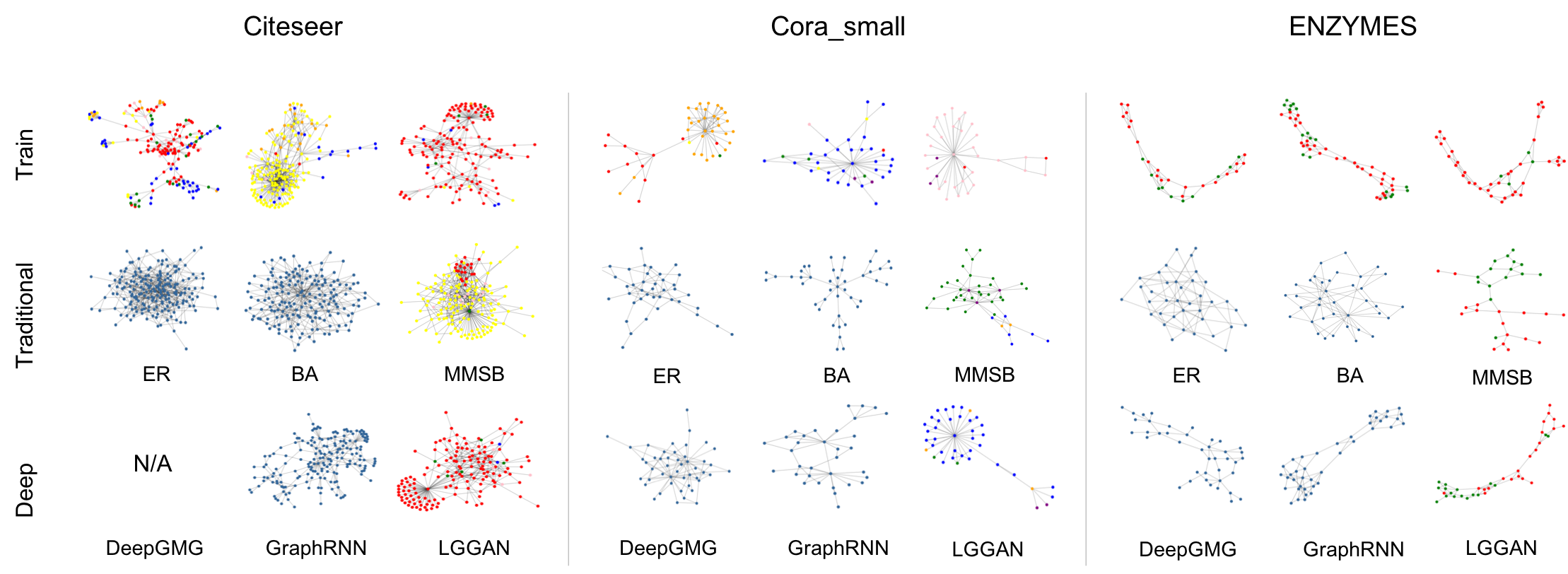}}
\caption{Visualization of training graphs (first row); graphs generated by traditional models (second row): E-R model, B-A model, and MMSB model; graphs generated by deep models (third row): DeepGMG, GraphRNN, and LGGAN for different datasets.}
\label{fig:gen_graphs}
\end{figure*}

\subsection{Generality} 

To evaluate the generality of these methods on graphs with different sizes, we perform experiments on two different subsets of the Cora dataset with different graph sizes: the Cora-small and Cora datasets. As listed in Table \ref{tab:evaluation}, the traditional models all create a large gap between these two datasets in terms of the three evaluation metrics. For the deep generative models, DeepGMG cannot generate large graphs due to the computational complexity of its generation procedure, which add nodes sequentially one at a time. And compared to GraphRNN, LGGAN MMD scores barely increase compared to smaller dataset, suggesting that our model is more reliable and has the best ability to scale up to large graphs.

Moreover, to evaluate the ability of LGGAN to adapt to different graph-structured data, we evaluate the results of all methods on the different domains of citation ego-networks (Cora) and molecular protein graphs (ENZYMES). As shown in Table \ref{tab:evaluation}, LGGAN achieves more consistent results on various datasets compared to other models, some of which suffer from the issue of specialization (e.g., MolGAN can only be used to generate specific or limited types of graph-structured data).

To show the comparison between the quality of graphs generated by our model and other baselines in terms of both reality and diversity, some examples are visualized in Figure \ref{fig:gen_graphs}, which contains graphs generated by our model and the compared methods. Although it is not as intuitive for humans to assess as, e.g., natural images, due to the complexity of graph structures, one can still see that LGGAN appears to capture the typical structures of datasets better than other models. Moreover, the visualization also provided extra information to augment the evaluation metrics for performance evaluation.

\subsection{Diversity }
\label{sec:hist}

Another important evaluation metric for generative models is diversity. Therefore, a good labeled-graph generative model should be able to generate diverse examples. Normally, two types of diversity are important during evaluation: (1) \textbf {Diversity among generated examples} would capture the natural variations in real graphs; and  (2) \textbf{Diversity compared to training examples} ensures that the generative model is doing more than exactly memorizing some training examples and outputting copies of them. Generative models should balance the need for generated outputs to be new graphs unseen during training with the need to retain important properties of the real data. 

Therefore, to investigate to what extent our model can maintain these types of diversity, we calculate the Weisfeiler-Lehman kernel value for both the training graphs and the generated graphs (by MMSB and LGGAN) and compute the kernel distance $d$ between any graphs $g_i$ and $g_j$ as 
\begin{equation}
    d_{ij} = \sqrt{K(g_i, g_i) + K(g_j, g_j) - 2 K(g_i, g_j)}.
\label{equ:dis_wl}
\end{equation}
We plot histograms of the minimum distances between each generated example and the training set in Figure \ref{fig:hist} for four datasets, Cora-small, CiteSeer-small, PROTEINS, and ENZYMES. In each plot, the left column shows the minimum distance of each training graph to any other graph; the middle column shows the minimum distance for each graph generated by LGGAN to the training graphs; and the right column shows the minimum distances for MMSB. These plots suggest that the graphs generated by our model are more similar to training graphs than the examples generated by MMSB, yet they are not exact copies of training graphs and have a similar diversity of graph distances as the real data based on the assumption that variations in distance also implies diversity in generated graphs.

\section{Conclusion}

In this work, we proposed a deep generative model using a GAN framework that generates labeled graphs. These labeled graphs can mimic distributions of citation graphs, knowledge graphs, social networks, and more. We also introduced an evaluation method for labeled graphs to measure how well the model learns the sub-structure of the labeled graphs. Our model can be used for simulation studies when access to labeled graph data is limited by privacy concerns. We can use these models to generate synthetic datasets or augment existing datasets to do graph-based analyses. Our experiments show that LGGAN outperforms other state-of-the-art models for generating graphs while also being capable of generating labels for nodes. The extra information in the node labeling task may help the model generate more realistic structures. In future work, we aim to address the common challenge across all graph generation methods of computational cost.


\bibliography{bigdata2020}
\bibliographystyle{IEEEtranSN}

\end{document}